\documentclass[runningheads]{llncs}
\usepackage[T1]{fontenc}
\usepackage{graphicx}
\usepackage{url}
\begin{document}
\title{Stain Based Contrastive Co-training for Histopathological Image Analysis}
%
\author{Bodong Zhang\inst{1,2}\orcidID{0000-0001-9815-0303} \and
Beatrice Knudsen\inst{3}\orcidID{0000-0002-7589-7591} \and
Deepika Sirohi\inst{3}\orcidID{0000-0002-0848-4172} \and
Alessandro Ferrero\inst{2}\orcidID{0000-0002-8671-1211} \and
Tolga Tasdizen\inst{1,2}\orcidID{0000-0001-6574-0366}}

\authorrunning{B. Zhang et al.}
\institute{Electrical and Computer Engineering, University of Utah, SLC, UT, USA\and
Scientific Computing and Imaging Institute, University of Utah, SLC, UT, USA\and
Huntsman Cancer Institute, University of Utah, SLC, UT, USA
\email{bodong.zhang@utah.edu,beatrice.knudsen@path.utah.edu,deepika.sirohi\\
@hsc.utah.edu,alessandro.ferrero@utah.edu,tolga@sci.utah.edu}
}
\maketitle              
\begin{abstract}
We propose a novel semi-supervised learning approach for classification of histopathology images. We employ strong supervision with patch-level annotations combined with a novel co-training loss to create a semi-supervised learning framework. Co-training relies on multiple conditionally independent and sufficient views of the data. 
We separate the hematoxylin and eosin channels in pathology images using color deconvolution to create two views of each slide that can partially fulfill these requirements. Two separate CNNs are used to embed the two views into a joint feature space. We use a contrastive loss between the views in this feature space to implement co-training. 
We evaluate our approach in clear cell renal cell and prostate carcinomas, and demonstrate improvement over state-of-the-art semi-supervised learning methods.

\keywords{Histopathology  \and Semi-supervised learning \and Co-training}
\end{abstract}
\section{Introduction}

Convolutional neural networks (CNNs) are commonly used in histopathology. Because digital whole slide images (WSIs) in pathology are much larger than typical input sizes for CNNs, workflows typically first tile the WSI into many smaller patches. There are two main approaches for training classification models with WSIs: strong and weak supervision. Strong supervision uses labels for the individual tiles, which requires expert annotation at a high cost~\cite{Dimitriou2019}. Weak supervision applies multiple instance learning with slide level labels~\cite{lerousseau:hal-03133239,Huang2019,Chikontwe2020,Campanella2019}. Weakly supervised methods have become popular due to the ease of obtaining labels for learning directly from pathology reports~\cite{Bulten:2022vv}. However, successful model training with weak learning requires thousands of WSIs, and strong supervision is still essential when a smaller number of WSIs are available for learning. 

Expert annotation at the tile level is infeasible to obtain beyond a small number of WSIs. Semi-supervised learning (SSL) seeks to leverage unlabeled data to improve the accuracy of models when only a limited amount of labeled data is available. One of the recent trends in SSL, consistency regularization~\cite{NIPS2016_30ef30b6,DBLP:conf/iclr/LaineA17}, has also found application in the classification of histopathology images. Teacher-student consistency~\cite{NIPS2017_68053af2} has been used to supplement tile-level labels for quantifying prognostic features in colorectal cancer~\cite{Shaw2020TeacherStudentCF} and in combination with weak supervision for Gleason grade classification in prostate cancer~\cite{Otalora2020}. The MixMatch model~\cite{NEURIPS2019_1cd138d0}
has been tested on histology datasets with open-set noise~\cite{9287980}. Weak/strong data transformation consistency (FixMatch)~\cite{NEURIPS2020_06964dce} has been applied to detection of dysplasia of the esophagus~\cite{Pulido2020}. State-of-the-art SSL methods rely on enforcing prediction/representation consistency between various transformations of the data. Whereas consistency under model perturbations has been proposed~\cite{NIPS2016_30ef30b6}, it is a less explored area. On the other hand, the co-training~\cite{Blum1998} approach to SSL can provide excellent results when multiple views of each sample are available that meet the criteria of sufficiency (each view should be able to support accurate classification on its own) and conditional independence given the label of a sample. 

Hematoxylin (H) and Eosin (E) are chemical stains that are used to highlight features of tissue architecture in formalin-fixed and paraffin-embedded tissue sections. 
H and E provide complementary information for pathologists. H is a basic chemical compound that binds negatively charged nucleotides in DNA and RNA to provide a blue color. In contrast, E is acidic and reacts with basic side chains of amino acids resulting in pink coloration. 
Whereas proteins bind to DNA and RNA lead to overlapping H and E staining in the cell nucleus and cytoplasm, the extracellular matrix and vascular structures supporting cancer cells interact primarily with E since they are devoid of DNA and RNA. In contrast to RGB channels that cannot easily be linked to a biological interpretation, H and E allow separation of nuclei versus cytoplasm and extracellular matrix. 
Therefore, we hypothesize that H and E stains, when separated into their own channels, can provide two views that can, to a large extent, satisfy the co-training assumptions.  We also formulate a novel contrastive co-training with H and E views. We validate our approach on a dataset of 53 WSIs from clear cell renal cell carcinoma (ccRCC) patients for histologic growth pattern (HGP) classification and of 45 WSIs from prostate cancer patients for cancer vs. benign gland classification. We demonstrate that our approach outperforms state-of-the-art SSL methods. We perform further experiments to explain the suitability of H and E channels for co-training as opposed to RGB channels. 

\section{Stain Based Contrastive Co-training}
\subsection{Stain Separation} 
We adopt an approach that separates an H\&E image in the RGB space into individual H and E stain channels using non-linear pixel-wise functions derived from dominant color profiles of each stain~\cite{Ruifrok2001QuantificationOH}. Concretely, we use the following approximate transformation between the two spaces:
\begin{equation}
    \left[ \begin{array}{c}
    H\\ E
    \end{array} \right] =
 \left[ \begin{array}{ccc}
    1.838 & 0.034 & -0.760\\ -1.373 & 0.772 & 1.215
    \end{array} \right]
    \left[ \begin{array}{c}
    \log _{10} 255/R\\
    \log _{10} 255/G\\
    \log _{10} 255/B\\
    \end{array} \right].
\end{equation}
The H and E channels are normalized to the range $[0,1]$ after the transformation. 
\begin{figure}
    \centering
    \includegraphics[width=0.67\textwidth]{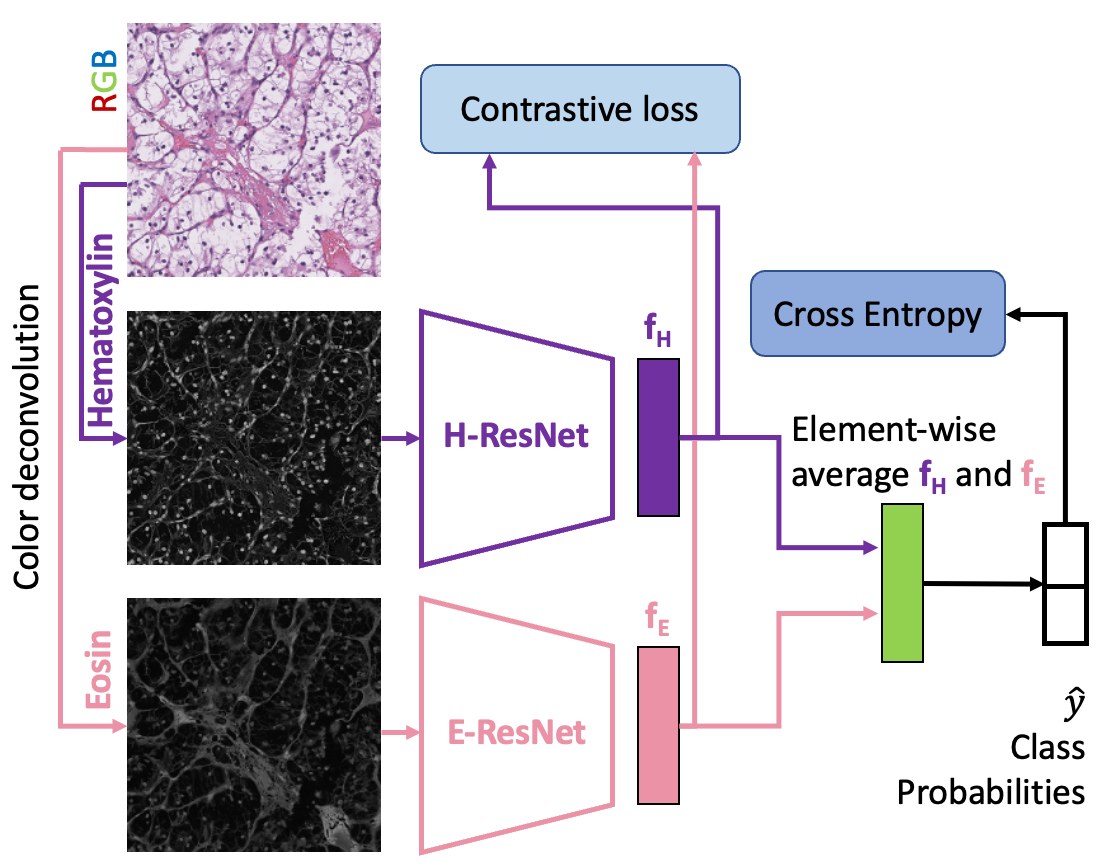}
    \caption{H\&E RGB image is separated into H and E channels and processed separately to generate two feature sets $f_H$ and $f_E$ trained with the proposed contrastive loss.}
    \label{fig:model}
\end{figure}

\subsection{Contrastive Co-training}
We propose two ResNet models~\cite{He2016DeepRL} (same architecture, separate parameters) for H and E channels, respectively (Figure~\ref{fig:model}). Existing co-training methods enforce consistency of prediction between classifier outputs operating on different views of the data. The disadvantage of this approach is that the individual classifiers only make use of their respective views and are sub-optimal. Instead, we propose a contrastive loss in the feature space to implement co-training and define a single classifier which uses a combined view by averaging the features from the two channels (Figure~\ref{fig:model}). Our approach is inspired by recent works that use contrastive learning to create a shared feature space between multimodal data~\cite{Yuan2021MultimodalCT,Yuhao2020}. We use a contrastive loss to create a shared feature space between features extracted by the H and E networks. Let $f_{H}(x)$ and $f_{E}(x)$ denote the H and E features  for input tile $x$, respectively. We use a triplet loss 
\begin{equation}
{\cal L}_{c.t.}(x_i)=\max\left(
\parallel f_H(x_i) - f_E(x_i)\parallel_2 - \parallel f_H(x_i) - f_E(x_k)\parallel_2 + m
,0\right),
\label{eqn:triplet}
\end{equation}
where random $k\neq i$, $\parallel a\parallel_2$ denotes the L2 norm of vector $a$, and $m$ is the margin hyperparameter. The triplet loss encourages $(f_H,f_E)$ pairs from the same H\&E tile $x_i$ to be mapped closer together than $(f_H,f_E)$ pairs from mismatched input tiles $x_i$ and $x_k$. Note that the output of the model is a linear+softmax layer applied to $0.5(f_H+f_E)$. Therefore, pushing the features $f_H$ and $f_E$ closer for the same tile also implicitly minimizes the difference between individual predictions, similar to co-training, if the final layer were applied to $f_H$ and $f_E$ alone. 

Let $L=\{x_j,y_j\}$ denote the labeled training set where $y_i$ is the label corresponding to input tile $x_i$. Let $U=\{x_i\}$ denote the unlabeled training set. 
The overall learning strategy combines supervised learning with cross-entropy on the labeled dataset with the triplet loss (\ref{eqn:triplet}) on the entire dataset: 
\begin{equation}
    {\cal L}=\sum_{x_j \in L} y_j\log \hat{y}_j
    + \lambda \sum_{x_i \in L\cup U} {\cal L}_{c.t.}(x_i),
    \label{eqn:loss}
\end{equation}
where $\hat{y}_j$ denotes the output of the model for input $x_j$ and $\lambda$ is a hyperparameter controlling the relative contributions from the labeled and unlabeled losses.  

\section{Experiments}
\subsection{Datasets}

\begin{figure}
    \centering
    \begin{tabular}{cc||cc}
    \includegraphics[height=0.21\textwidth]{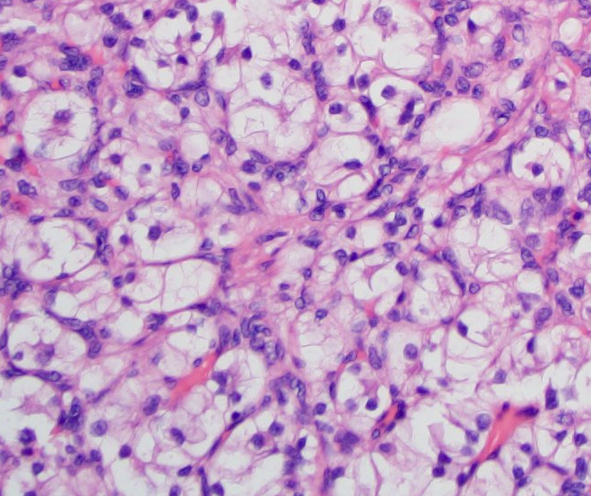}&
    \includegraphics[height=0.21\textwidth]{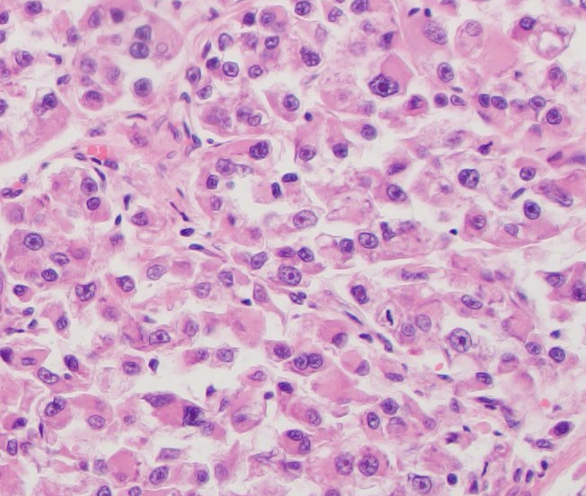} &
    \includegraphics[height=0.21\textwidth]{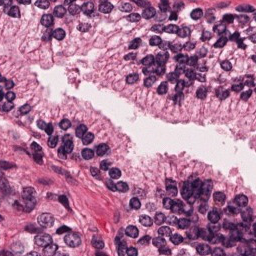}&
    \includegraphics[height=0.21\textwidth]{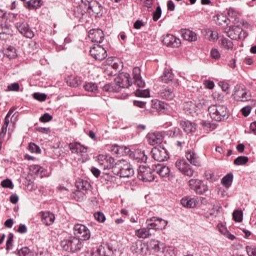}\\
        ccRCC nested & ccRCC diffuse &
    Prostate benign  & Prostate cancer
    \end{tabular}
    \caption{Examples of ccRCC and prostate gland tiles. 
    }
    \label{fig:tiles_examples}
\end{figure}

\noindent {\bf ccRCC.}
H\&E slides from ccRCC patients at our institution were retrieved from the pathology archive and scanned 
at $40\times$ magnification. 
HGPs in 53 WSIs were annotated by drawing polygons around them in QuIP~\cite{Saltz2017} by a GU-subspecialty trained pathologist. HGPs were divided into nested vs. diffuse (non-nested) histologic classes~\cite{Sirohi2022}. Diffuse HGPs are associated with a higher risk of cancer recurrence and metastatic progression. Each WSI contains multiple polygons. Images were downsampled by a factor of $2\times$ and a tile size of $400\times 400$ pixels was chosen to capture the visual characteristics of the HGPs after discussion with pathologists (Figure~\ref{fig:tiles_examples}). We sampled overlapping tiles by choosing a stride of $200$ pixels. We extracted 3014 tiles with nested HGPs and 2566 tiles with diffuse HGPs from the annotated polygons. We separated the WSIs into training, validation and testing sets to ensure a realistic experimental setting. This separation resulted in 2116/1990 nested/diffuse tiles for training, 386/246 nested/diffuse tiles for validation and 512/330 nested/diffuse tiles for testing. The validation set was used for choosing hyperparameters as discussed below. The test set was for final model evaluation. Tiles from same patient were in the same set. For the SSL experiments, we divided the annotated polygons from the training set into 10 groups and randomly picked one group to draw labeled tiles from in each run of our experiments. This is a more realistic and challenging scenario than randomly choosing $10\%$ of the training tiles as labeled data because tiles from the same polygon usually represent a smaller range of variations for learning.

\noindent {\bf Prostate Cancer.} 
We collected 6992 benign gland images and 6992 prostate cancer images from our institution as training set using the same process as with the ccRCC dataset. 
The tile size was chosen as $256\times 256$, which is sufficient to characterize gland features. The validation and test sets are from the The Cancer Genome Atlas Program (TCGA). We collected 477 benign and 472 cancer tiles from 18 cases. In each experiment, we randomly selected 8 cases for validation and 10 cases for testing. Examples of prostate gland images are shown in Figure~\ref{fig:tiles_examples}. 
For SSL, we randomly divided training images into 20 groups and used one group as labeled data. The image sets from our institution are available through a material transfer agreement and the TCGA set is publicly available.

\subsection{Model Selection, Training and Hyperparameters}
Considering the small number of training samples, we chose ImageNet pretrained ResNet18~\cite{He2016DeepRL} for all experiments. ResNet is a state-of-the-art model which has better performance with less parameters. For models that use single channel inputs, i.e., the H and E CNN pathways in Figure~\ref{fig:model}, we summed the convolutional weights of the R, G and B channels in the first layer of ResNet18. The final layer of the ResNet18 was also changed for binary classification.

We used color jittering, random rotation, crop to $256\times 256$(ccRCC) or $224\times 224$(prostate) pixels, random horizontal/vertical flip and color normalization as data augmentation. For validation and test tiles, we performed center crop and color normalization to follow the same data format as in training. For the co-training model, H and E channels have independent color jitters but the rest of the augmentations are common, e.g., the same random rotation angle is applied to the H and E channels from the same tile. The rationale for independent color jitters is that color variations due to the amount of H or E chemical tissue stains used are common in practice, which leads to independent brightness variations in these channels. 

 We used the Adam optimizer with an initial learning rate of $10^{-3}$(100\% label only) or $10^{-4}$ and used a decaying learning rate. 
 A batch size of $64$ was used in ccRCC dataset and 128 in prostate cancer dataset. Hyperparameters in (\ref{eqn:loss}) were chosen as $\lambda=0.2\times p$ and $m=40$, where $p$ is the percentage of training data used as labeled data. All hyperparameters were chosen to optimize accuracy over the validation set, including experiments on other state-of-the-art models. Batch normalization was applied to the features before computing the contrastive loss. 

For comparison with other state-of-the-art SSL methods, we used consistency regularization~\cite{NIPS2016_30ef30b6,DBLP:conf/iclr/LaineA17}, MixMatch~\cite{NEURIPS2019_1cd138d0} and FixMatch~\cite{NEURIPS2020_06964dce}. 
 The same augmentations discussed above were used for the SSL experiments. 
We ran all experiments for 250 epochs in ccRCC experiments and 100 epochs in prostate cancer dataset and chose the epoch with the best validation accuracy. 
Each experimental setting was run 5 times to calculate mean accuracy and standard deviation.

We used Python 3.7.11 $+$ Pytorch 1.9.0 $+$ torchvision 0.10.0 $+$ CUDA 10.2 on virtual environment and ran on NVIDIA TITAN X and NVIDIA TITAN RTX. We also used Python 3.9.0 $+$ Pytorch 1.7.1 $+$ torchvision 0.8.2 $+$ CUDA 11.0 and ran on NVIDIA RTX A6000. With batchsize fixed to 64, co-training experiment on ccRCC occupied around 5300MB memory on GPU and needed 1.5-2.0 minutes for each epoch. The code is available at \url{https://github.com/BzhangURU/Paper_2022_Co-training}\par

\subsection{Results}
We compared proposed co-training with H and E views to a baseline ResNet18 model that uses RGB H\&E images as input, as well as other state-of-the-art SSL methods, such as MixMatch and FixMatch, considering they are already widely used in histopathology image analysis~\cite{9287980}. 
The approaches were compared under two settings: using 100\% of the available labeled tiles in training set for supervised learning and using only a subset (10\% in ccRCC, 5\% in prostate) of the available tiles for supervised learning. The proposed model also employed the unsupervised co-training loss with 100\% of the training data (unlabeled) to set up an SSL method. 
Mean accuracy and standard deviation over 5 runs reported for all methods are shown in Table~\ref{ref:tab} for both datasets. 

\begin{table}
\begin{center}
\caption{Mean accuracy and standard deviations of different models for the test sets in ccRCC and prostate experiments. Best performing model results for the 100\% and 10\%/5\% labeled data setting are shown in bold.}
\label{ref:tab}
\begin{tabular}{|l|c||l|c|}
\hline
ccRCC Model &  Test Accuracy & Prostate Model & Test Accuracy\\
\hline
100\% label RGB ResNet &  $84.8\pm 2.4\%$ & 100\% label RGB ResNet &  $77.5\pm 2.5\%$ \\
100\% label H/E co-train & $\mathbf{92.0\pm 2.6\%}$ & 100\% label H/E co-train & $\mathbf{79.1\pm 2.0\%}$\\
\hline
10\% label RGB ResNet  &  $76.9\pm 5.9\%$ & 5\% label RGB ResNet  & $73.4\pm 1.0\%$ \\
10\% label RGB consis & $86.8\pm 3.3\%$ & 5\% label RGB consis & $74.7\pm 1.3\%$\\
10\% label RGB MixMatch & $85.9\pm 5.7\%$ & 5\% label RGB MixMatch & $73.7\pm 5.0\%$\\
10\% label RGB FixMatch & $88.3\pm 3.8\%$ & 5\% label RGB FixMatch & $78.2\pm 3.8\%$\\
10\% label H/E co-train & $\mathbf{92.3 \pm 2.1\%}$ & 5\% label H/E co-train & $\mathbf{78.7 \pm 1.9\%}$\\
\hline
\end{tabular}
\end{center}
\end{table}

 We note that the contrastive co-training strategy improved test accuracy, by a large margin in the case of ccRCC, when 100\% of the labeled data were used for supervised training (row 2 vs. 1, Table~\ref{ref:tab}), which suggests it provides a strong regularization effect against overfitting. Note that training accuracy for the fully supervised RGB ResNet and H/E co-train models were $99.97\pm 0.02\%$ and $99.78\pm 0.15\%$, respectively, in the ccRCC dataset. The same models achieve $98.34\pm 0.64\%$ and $98.32\pm 0.55\%$ training accuracy in prostate cancer. The fact that test accuracies on prostate cancer are lower than on ccRCC for all models is likely due to domain shift. In ccRCC dataset, all training, validation and test sets come from our institution. While in prostate cancer dataset, only training set comes our institution, the validation and test set come from TCGA dataset. Another possible reason is the fact that sometimes the gland size in prostate cancer is much smaller than the size of tiles, which could carry much less distinguishable features.

 As expected, the proposed co-training strategy significantly outperforms the baseline approach (row 7 vs. 3, Table~\ref{ref:tab}) under the limited labeled data setting. Consistency regularization based SSL methods significantly improve the accuracy of RGB ResNet baseline when a limited amount of training data is available (rows 4-6 vs. 3, Table~\ref{ref:tab}). In line with results from computer vision, FixMatch even surpasses the baseline model trained with the entire labeled dataset. However, our proposed method outperformed all other SSL methods we compared against including FixMatch for both datasets. We note that hyperparameters for all SSL methods were independently fine-tuned to obtain the best validation accuracy. Finally, contrastive co-training was able to reach the same accuracy levels independent of the amount of labeled data that was used for supervised training (rows 2 and 7, Table~\ref{ref:tab}).



\subsection{Co-training View Analysis}
\label{sec:views}
In this section, we further study the suitability of the H and E channels for co-training in the context of the ccRCC dataset. First, we explore whether the H and E channels are sufficient on their own to provide a basis for accurate classification in a supervised setting. We train models that only use the H or only use the E channel as input. The 100\% labeled results in Table~\ref{ref:table_of_H_only_E_only} show that both channels carry sufficient information for the classification problem at hand. This is especially true for the E-channel, which is particularly informative for the nested vs diffuse classification task. However, as expected, the accuracy for both channels drops significantly when the labeled data is limited. 

\begin{table}
\begin{center}
\caption{H-only and E-only models test accuracy for the ccRCC dataset.}
\setlength{\tabcolsep}{3mm}
\label{ref:table_of_H_only_E_only}
\begin{tabular}{|l|c|l|c|}
\hline
Model &  Accuracy &
Model &  Accuracy\\
\hline
100\% label H ResNet &  $79.4\pm 3.7\%$ & 
10\% label H ResNet &  $73.5\pm 4.0\%$\\
\hline
100\% label E ResNet &  $94.0\pm 1.4\%$&
10\% label E ResNet &  $82.3\pm 7.0\%$\\
\hline
\end{tabular}
\end{center}
\end{table}

We next explore if the H and E channels are better suited for co-training than R, G and B channels due to a higher degree of independence. We trained an image-to-image regression model using the U-Net architecture~\cite{Ronneberger2015} between various channels, e.g., predicting the E-channel from the H-channel of the same tile. The final layer of the U-Net architecture was chosen to be linear, and we used the mean square error function for training. Table~\ref{ref:table_of_choosing_views} reports the coefficient of determination ($R^2$) achieved for various input/output channel combinations. We observe that the H and E channels are harder to predict from each other (lower $R^2$) compared to the R, G and B channels, hence demonstrating a higher degree of independence and suitability for co-training.

\begin{table}
\begin{center}
\caption{Coefficient of determination($R^2$) of image mapping between various channels on ccRCC validation set at epoch with the lowest MSE. 
}
\setlength{\tabcolsep}{3mm}
\label{ref:table_of_choosing_views}
\begin{tabular}{|l|c|l|c|}
\hline
Experiments &  $R^2$ value & Experiments &  $R^2$ value\\
\hline
H $\Rightarrow$ E &  $0.5223$ &
E $\Rightarrow$ H &  $0.4613$ \\
\hline
R $\Rightarrow$ G & $0.8464$ &
G $\Rightarrow$ R & $0.7833$\\
\hline
R $\Rightarrow$ B & $0.8207$ &
B $\Rightarrow$ R & $0.7713$ \\
\hline
G $\Rightarrow$ B & $0.8522$ &
B $\Rightarrow$ G & $0.8824$ \\
\hline
\end{tabular}
\end{center}
\end{table}


\subsection{Ablation Studies}
We also conducted ablation studies to separately analyze the role of the contrastive loss and the H and E channel selection in terms of classification accuracy on ccRCC. 
Omitting the contrastive loss from training while using the H and E channel inputs lowered the accuracy from $92.0\pm 2.6\%$ to $84.7\pm 5.2\%$ for 100\% labeled data and from $92.3 \pm 2.1\%$ to $78.7\pm 8.0\%$ for 10\% labeled data. In the next ablation experiment, we used various pairs from the RGB channels as the basis for our co-training method, and compared with ResNet using the same pair as input, e.g., using only the R and B channels to form 2-channel images as input for ResNet. Results are reported in Table~\ref{ref:tab_ablation}. Unlike the H and E models, we observe that the results are approximately the same, which is expected considering the higher level of dependence among RGB channels shown in Section~\ref{sec:views}.
These observations suggest that the benefit of the proposed model is due to the contrastive co-training loss applied to the H and E view inputs rather than simply due to the change in the input space or the contrastive loss individually.  
\begin{table}
\begin{center}
\caption{Ablation study on ccRCC. Test set accuracy of ResNet and co-training models taking only 2 channels from RGB as input with 10\% labeled data in training.}
\label{ref:tab_ablation}
\begin{tabular}{|l|c||l|c||l|c|}
\hline
Model &  Accuracy &
Model &  Accuracy &
Model &  Accuracy \\
\hline
RB ResNet & $77.5\pm 6.6\%$ & RG ResNet & $80.2\pm 6.4\%$ &
GB ResNet & $78.4\pm 9.7\%$\\
\hline
R/B co-train & $78.2\pm 4.5\%$&
R/G co-train & $79.8\pm 5.6\%$&
G/B co-train & $76.6\pm 7.3\%$\\
\hline
\end{tabular}
\end{center}
\end{table}

\section{Conclusion}
We proposed a novel co-training approach for pathology image classification that leverages deconvolution of an H\&E image into individual H and E stains. We demonstrated the advantages of the proposed approach over fully supervised learning and other state-of-the-art SSL methods in the context of ccRCC and prostate cancer. 
The proposed method could be used after segmentation of cancer regions from a WSI to drive prognostic markers. In future work, we will investigate finer-grained classification for further improved prognostic value. 

 
 Since our proposed approach uses a complementary learning strategy to consistency regularization approaches that use data transformations, a potential avenue for future research is to combine them for further improvements.
 Another  potential direction for further research is to investigate whether a more sophisticated separation into H and E stain channels can provide improved results with co-training. Methods based on Cycle-GAN have been used for stain-to-stain translation such as H\&E to immunohistochemistry and they could also be used for separation of H and E stains. However, this would require the acquisition of additional datasets with H only and E only stains for the discriminators.

\noindent {\bf Acknowledgements.}
We are grateful for the support of the Computational Oncology Research Initiative (CORI) at the Huntsman Cancer Institute, University of Utah. We also acknowledge support of ARUP Laboratories and the Department of Pathology at University of Utah.

%
%
%
%
\bibliographystyle{splncs04}
\bibliography{pathology,ML}

\end{document}